\documentclass{article}
\usepackage{ijcai23}

\usepackage{times}
\usepackage{soul}
\usepackage{url}
\usepackage[hidelinks]{hyperref}
\usepackage[utf8]{inputenc}
\usepackage[small]{caption}
\usepackage[table]{xcolor} 
\usepackage{graphicx}
\usepackage{amsmath}
\usepackage{amsthm}
\usepackage{booktabs}
\usepackage{algorithm}
\usepackage{algorithmic}
\usepackage[switch]{lineno}

\newcommand{\EVLP}{Embodied Vision-Language Planning}

\newcommand{\evlp}{EVLP}

\title{Core Challenges in \EVLP~(Extended Abstract)\footnote{The full version was published at JAIR \href{https://dl.acm.org/doi/pdf/10.1613/jair.1.13646}{[Francis \textit{et al.}, 2022]}.}}

\author{
    Jonathan Francis$^{1,2}$\footnote{Contact author.}, Nariaki Kitamura$^{3}$, Felix Labelle$^{2}$, Xiaopeng Lu$^{2}$, Ingrid Navarro$^2$, Jean Oh$^{2}$
    \affiliations
    $^{1}$Bosch Center for Artificial Intelligence\\
    $^{2}$School of Computer Science, Carnegie Mellon University\\
    $^{3}$Komatsu Limited
    \emails
    \{jmf1, ingridn, jeanoh\}@cs.cmu.edu, nariaki\_kitamura@global.komatsu, \\\{flabelle, xiaopen2\}@alumni.cmu.edu 
}

\begin{document}

\maketitle

\newcommand{\jf}[1]{\textcolor{red}{[\textbf{JF:} #1]}}
\newcommand{\fl}[1]{\textcolor{blue}{[\textbf{FL:} #1]}}
\newcommand{\jo}[1]{\textcolor{magenta}{[\textbf{JO:} #1]}}
\newcommand{\na}[1]{\textcolor{cyan}{[\textbf{IN:} #1]}}
\newcommand{\nav}[1]{\textcolor{cyan}{[\textbf{IN:} #1]}}
\newcommand{\xl}[1]{\textcolor{orange}{[\textbf{XL:} #1]}}
\newcommand{\nk}[1]{\textcolor{green}{[\textbf{NK:} #1]}}
\newcommand{\parentcite}[1]{\cite{#1}}%
\newcommand{\parencite}[1]{\cite{#1}}%
\newcommand{\textcite}[1]{\cite{#1}}%
\newcommand{\para}[1]{{\noindent\textbf{#1}}}
\newcommand{\comment}[1]{}
\newcommand{\cut}[1]{}
\newcommand{\link}[1]{\noindent\textcolor{red}{\textbf{#1}}}

\begin{abstract}
Recent advances in the areas of Multimodal Machine Learning and Artificial Intelligence (AI) have led to the development of challenging tasks at the intersection of Computer Vision, Natural Language Processing, and Robotics. Whereas many approaches and previous survey pursuits have characterised one or two of these dimensions, there has not been a holistic analysis at the center of all three. Moreover, even when combinations of these topics are considered, more focus is placed on describing, e.g., current architectural methods, as opposed to \textit{also} illustrating high-level challenges and opportunities for the field. In this survey paper, we discuss \EVLP~(\evlp) tasks, a family of prominent embodied navigation and manipulation problems that jointly leverage computer vision and natural language for interaction in physical environments. We propose a taxonomy to unify these tasks and provide an in-depth analysis and comparison of the current and new algorithmic approaches, metrics, simulators, and datasets used for \evlp~tasks. Finally, we present the core challenges that we believe new \evlp~works should seek to address, and we advocate for task construction that enables model generalisability and furthers real-world deployment. 
\end{abstract}

\section{Introduction}

\begin{figure*}[!tp]
    \centering
    \includegraphics[width=\textwidth,keepaspectratio]{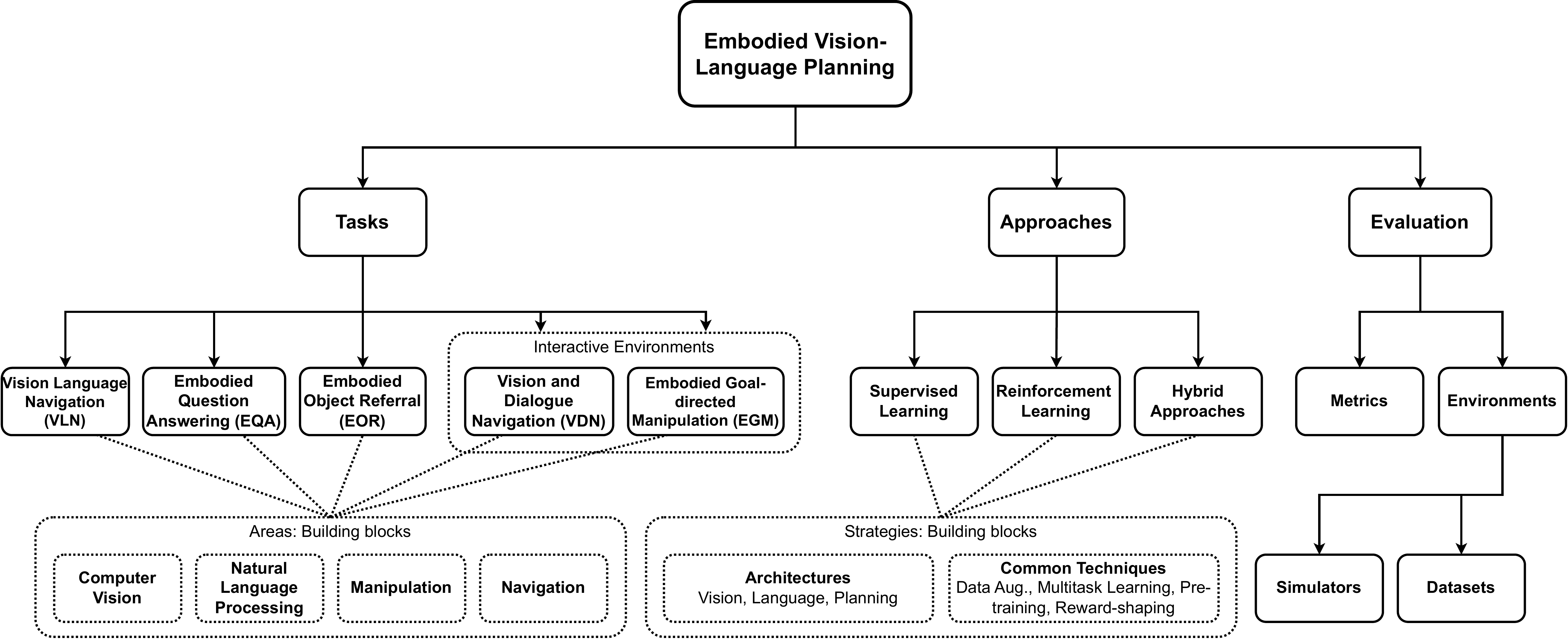}
    \caption{Taxonomy of \textit{\EVLP}.} % (Don't delete) Figure edit link: https://app.diagrams.net/#G1p8RFCQIN_T1H8_CfxdHnbE5dIYCaxKJQ
    \label{fig:taxonomy}
    \vspace{-0.4cm}
\end{figure*}

With recent progress in the fields of Artificial Intelligence (AI) and Robotics, intelligent agents are envisaged to interact with humans in shared environments. Such agents include any entities that can make decisions and take actions autonomously and are expected to understand semantic concepts in those environments, using, e.g., visual, haptic, auditory, or textual information perceived via sensors \parencite{wooldridge1995intelligent,castelfranchi1998modelling}. With the goal of developing intelligent agents equipped with these sensory and reasoning capabilities, Embodied AI (EAI), as a field, has become popular for studying the particular set of AI problems surrounding agents situated in a physical environment: recently, the number of papers and datasets for the tasks that require the agents to use both vision and language understanding has increased markedly~\cite{EQA,IQA,R2R,VLN-CE,CVDN,nguyen2019hanna,VLNBERT,FineGrainedR2R}. In this article, we conduct a survey of recent works on these types of problems, which we refer to as \textit{\EVLP}~(\evlp)~tasks. In this article, we aim to provide a bird's-eye view of current research on \evlp~problems, addressing their main challenges and future directions. Our main contributions are as follows: (i) We formally define the field of \EVLP~and we propose a taxonomy that both unifies a set of related tasks in \evlp~and serves as a basis for categorising new tasks; (ii) we survey recent \evlp~tasks, compare their task properties, highlight modelling approaches used in those tasks, and analyse the datasets, simulators, and metrics used to evaluate the approaches on those tasks; finally, (iii) we identify open challenges that afflict existing works in the \evlp~family, with an emphasis towards encouraging unseen generalisation and deploying algorithms to the real world. We refer readers to our full journal article for further details \cite{francis2022core}.

\subsection{Problem Definition}

We discuss a broad set of problems, related to an embodied agent’s ability to make planning decisions in physical environments. Formally, let $S$ and $A$ denote sets of states and actions; $V$ and $L$ denote sets of vision and language inputs available to the agent. A planning problem is defined by the tuple $\Phi = \{S, A, s_{ini}, s_{goal}\}$, where $s_{ini}, s_{goal} \in S$ denote initial and goal states, respectively. 
A solution $\psi \in \Psi_{\Phi}$ to planning problem $\Phi$ is a sequence of actions to take in each state, starting from an initial state to reach a goal state, $\psi = [s_{ini}, a_0, ..., s_t, a_t, ..., a_T, s_{goal}]$, where $t\in T$ is a finite time-step in episode length $T$ and $\Psi_{\Phi}$ is a set of possible solutions to $\Phi$. Given a particular \evlp~problem $\Phi$, state $s_t \in S$ at time step $t$ can be defined in terms of vision and language inputs up to the current time step, such that, $s_t = \{(v_0, l_0), (v_1,l_1), \dots, (v_t,l_t), \dots, (v_T,l_T)\}$, where $v_{t} \in V$ and $l_{t} \in L$. The agent's objective is to minimize the difference between an admissible solution $\psi \in \Psi_{\Phi}$ and its predicted one $\bar{\psi}$. This definition broadly captures the crux of \evlp~problems. Customized definitions are needed for specific tasks, where additional constraints or assumptions are added to focus on particular subareas of this general problem.

\subsection{Taxonomy}

We propose a taxonomy of \evlp~research, illustrated in \autoref{fig:taxonomy}, around which the rest of the paper is organized. The taxonomy subdivides the field into three branches; tasks, approaches, and evaluation methods. The Tasks branch proposes a framework to classify existing tasks and to serve as a basis for distinguishing new ones. The Approaches branch touches on the learning paradigms, common architectures used for the different tasks, as well as common tricks used to improve performance. The right-most branch of the taxonomy discusses task Evaluation Methodology, which is subdivided into two parts: metrics and environments. The metrics subsection references many of the common metrics, used throughout \evlp~tasks, while the environments subsection presents the different simulators and datasets currently used.

\section{Current Approaches}

We provide a brief overview of the tasks, methodology, learning paradigms, datasets, simulators, and metrics used in the \evlp~task family; we provide additional references, task-specific problem definitions, architectural descriptions and training objectives, dataset and simulator comparisons and statistics, and metric formul\ae~in \cite{francis2022core}.

\subsection{Tasks, Methods, and Learning Paradigms}

\para{\evlp~Tasks.} Many \evlp~tasks have been proposed, with each task focusing on different technical challenges and reasoning requirements for agents. Tasks vary on the basis of the action space (types and number of actions possible), the reasoning modes required (e.g., instruction-following, versus exploration and information-gathering), and whether or not the task requires interaction with another agent. Vision-Language Navigation and Vision and Dialogue Navigation require agents to use natural language instructions to navigate to goal locations in environments, where the latter provides agents with intermediate supervision and clarifications. In Embodied Question Answering tasks, an agent initially receives a language-based question, and must engage in guided exploration of the environment, in order to collect enough information about its surroundings to generate an answer. In Embodied Object Referral tasks, an agent navigates to an object mentioned in a given instruction, and has to identify (or select) it upon reaching its location. Embodied Goal-directed Manipulation tasks combine manipulation-based environment interactions with requirements from aforementioned tasks, such as navigation and path-planning, state-tracking, instruction-following, instruction decomposition, and object-selection.

\para{Methods and Learning Paradigms.} Technical approaches that pursue solutions to \evlp~tasks must model various facets. Firstly, modelling vision typically involves building a lossless and predictive state representation of the agent's environment; because ego-centric visual observations change as the agent navigates or manipulates objects in the space, the agent must also include temporal modelling mechanisms, in order to represent observed state-changes in its environment over time and to monitor progress of task-execution. Next, modelling language in \evlp~tasks typically requires using the instructions or question prompts provided to generate a rich description of the agent's goal; because language can be ambiguous in practice, challenges remain in obtaining unbiased representations. Next, the agent must be able to compare its progress in the task with its representation of the goal, typically requiring sophisticated strategies for multimodal representation, alignment, and fusion. Finally, in order to interact with their environments, agents must include mechanisms for action-generation and planning; inspired by classical approaches in robotics, many such approaches follow from early works in mapping and exploration strategies, search and topological planning, and hierarchical task decomposition. To bias agents towards desired task-oriented behaviour, approaches leverage various learning paradigms (e.g., semi/self/fully-supervised learning, reinforcement learning, etc.) and strategies (e.g., pre-training, data augmentation, multitask learning, reward-shaping, cycle-consistency, etc.).

\subsection{Datasets, Simulators, and Metrics}

\para{Datasets.} \evlp~datasets vary across three primary main dimensions: visual observations, natural language inputs, and expert demonstrations. Visual observations, in general, consist of RGB images often paired with depth data or semantic masks. These observations can represent both indoor and outdoor environments from both, photo-realistic or synthetic-based settings. In contrast, language varies in the type of prompt. Language prompts may come in the form of questions, step-by-step instructions, or ambiguous instructions that require some type of clarification through dialog or description. Language can also vary in terms of complexity of language sequences and scope of vocabulary. Finally, navigation traces differ in aspects like the granularity (or discretization) of the action-space and the implicit alignment that a provided action sequence has with the other two dimensions. 

\para{Simulators.} Early simulation platforms for EAI research typically leveraged simple video game environments to create and train neural controllers. Human performance was quickly achieved on many of these platforms, as simplified environments generally lack the diversity and complexity of real-world settings. Recent works have addressed this lack of realism through the use of photo-realism and the use of interactive contexts where agents are able to modify the states of objects in the environment. Toward this end, there is also interest in developing frameworks focused on simulation-to-real transfer and evaluation, allowing the study of discrepancies between real settings and simulated ones. Finally, other platforms have also focused on encouraging reproducibility of work, flexibility of design, and benchmarking.

\begin{figure*}[!tp]
    \centering
    \includegraphics[width=\textwidth,keepaspectratio]{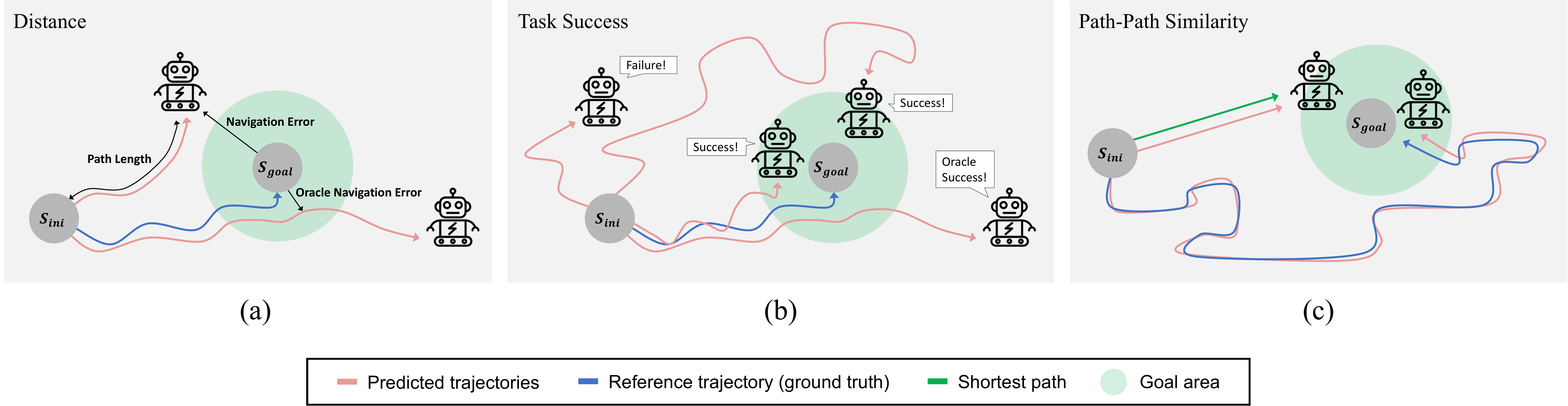}
    \caption{Phenomena measured by typical metrics used in \textit{\EVLP} tasks.}
    \label{fig:metrics}
\end{figure*}

\para{Metrics.} Popular metrics in \evlp~research can be grouped into categories---each measuring a different aspect of agent performance, such as distance (quantifies the manner in which an agent traversed a space), success (characterises extent to which the overall task is completed by an agent), path-path similarity (assesses the extent to which the agent's trajectory was similar to the ground-truth), instruction-based metrics (measures the alignment between natural language instructions and the agent's trajectory), and object-centric metrics (assess efficacy of object selection, identification, or manipulation). We illustrate the first three, in Figure \ref{fig:metrics}.

%\para{Metrics.} Popular metrics used in \evlp~research can be grouped into categories, each measuring a different aspect of agent performance, such as distance, task success, path-path similarity, instruction-based metrics, and object-centric metrics. We illustrate the first three, relating particularly to the assessment of agent behaviour, in Figure \ref{fig:metrics}. Distance metrics seek to quantify the manner in which an agent traversed a space. Examples include path length (total distance traveled), navigation error (displacement between the agent's final state and the goal location), and oracle navigation error (displacement between the goal location and the closest point on the agent's executed trajectory). Success metrics seek to characterise the extent to which the overall task is completed by an agent--e.g., success rate (how often the agent arrives within a threshold distance of the goal) and success weighted by average inverse path length. Finally, path similarity metrics assess the extent to which the agent's trajectory was similar to the ground-truth. 

\section{Core Challenges}
\label{sec:core_challenges}

\subsection{New Directions in EVLP Research}
We highlight three promising directions, in the pursuit of more ubiquitous human-robot interaction and better agent generalisation. Firstly, we advocate for improved \textit{social interaction}: we feel that a progression from static instructions to active dialogues would enable new collaborative and assistive capabilities to emerge. Next, to enable agents that accommodate more complexities of real-world deployment, we advocate for the introduction of \textit{dynamic environments} in \evlp~research, encouraging agents to incorporate reasoning strategies that are robust to environment uncertainty and non-stationarity. We discuss a vision for \textit{cross-task robot learning}, wherein agents may acquire experience from related modality-centric tasks, before their deployment to shared multimodal settings with significant task overlap. Finally, we would highlight new directions in interactive object perception for transfer learning, where agents must physically interact with the environment in order to learn new concepts \cite{tatiya2022transferring,tatiya2023cross}.

\subsection{Use of Domain Knowledge}
We further encourage the development of methods that utilise domain knowledge in a principled way, for guiding the learning and transfer of models; while this notion has seen a recent resurgence in other fields \cite{francis2022knowledge,park2020diverse,francis2022l2rarcv,herman2021learn,andreas2016neuralmodulenetworks,francis2019occutherm}, we notice few such works in \evlp. Domain knowledge comes in many forms, e.g., graphical models, logical rules, constraints, pre-training, knowledge graphs, and others; and while domain knowledge holds the promise of improving agents' sample-efficiency, interpretability, safety, and generalisability, the challenge exists in how to effectively express and utilise this domain knowledge in an arbitrary learning problem. \textit{Pre-training} and \textit{commonsense knowledge}, in particular, serve as two manifestations that show promise for imbuing agents with the aforementioned attributes. 

Pre-training tasks have been carefully designed and coupled with popular high-capacity models, for self-supervision in such domains as image classification \cite{Resnet} and natural language processing \cite{devlin2018bert,yang2019xlnet,ma2020zeroshotqa}, in attempts to maximise the generalisability of transferred or fine-tuned approaches. While there is some progress in the context of specific multimodal problems \cite{VLNBERT,PREVALENT,lu2019vilbert}, challenges remain for developing generalisable pre-training strategies that encompass the scope of the broader \evlp~task family.

Commonsense knowledge acquisition and injection in models remains an active research area in NLP~\cite{talmor2018commonsenseqa,ma2019generalizable,ma2020zeroshotqa,li2020lexicallyconstrained}, with some works proposing to ground observations with structured commonsense knowledge bases, directly, thereby improving downstream performance on relevant tasks. However, the use of commonsense knowledge in the context of \evlp~tasks remains largely unexplored. As the ultimate goal of \evlp~tasks is to develop intelligent agents that are capable of solving real-world problems in realistic environments, it is reasonable to consider providing models with structured external knowledge of the world 
\cite{VSNScenePriors,tatiya2022knowledge}.

\subsection{Agent Training Objectives}
Selecting the appropriate training objective(s) for agents undertaking a given task has been a long-standing problem in machine learning and artificial intelligence; this selection depends on the nature of the available training signals (e.g., reward or cost functions, level of supervision, environment observability) and on the degree to which external knowledge (e.g., auxiliary objectives, common sense, constraints) are deemed necessary for effectively biasing agent behaviour. For \evlp~tasks, the selection of training objectives is made more challenging by the complex nature of the environments in which agents operate. This often necessitates frameworks that consist of more than one biasing strategy. Given the underlying motivation of optimising for generalisability and interpretability, explicit treatment should be given to finding the learning paradigm(s) that most effectively integrate information for various related sources and generalises agents' inductive biases to new environments; indeed, the training paradigms should include explicit mechanisms for encouraging the properties we hope to imbue.

\subsection{Simulation-to-Real Gap}
Datasets and simulation environments are the primary driving forces behind \evlp~research, since the measure of model efficacy relies on the availability of strong testing scenarios and the appropriate evaluation criteria. Current \evlp~tasks are implemented as a set of goals and metrics, atop pre-existing simulators or datasets. In this section, we urge the community to consider and prioritise the deployment of \evlp~agents to real-world settings. Specifically, we assert that various \evlp~tasks and metrics may be improved on the basis of three dimensions: \textit{simulator realness}, \textit{dataset realness}, and \textit{tests for model generalisability}. 

Simulation-based training and execution are especially attractive when modelling a sequential learning problem, since offline datasets do not allow for such recursive interaction with an environment. There are limitations, however, in how effectively scientists and practitioners can encourage the desired model behaviour to emerge for real-world use-cases. Because this dissonance reduces models' immediate viability for real-world deployment, we assert the importance of increased attention from the computer vision and robotics communities on the topics of simulation-to-real transfer, unseen generalisation, robustness to out-of-distribution settings. We encourage the definition of metrics that assess intermediate agent behaviours and task efficiency, as opposed to simply indicating in-domain task completion.

Dataset-based training can be highly effective, e.g., when providing models with strong priors on agent behaviour. Relying solely on datasets for training can present significant issues, however, not least of all causal confusion, limited observation of the environment's transition dynamics, and unrealistic priors due to class imbalance. Additionally, tasks metrics that have been defined on top of datasets may vary in their ability to truly assess whether agents are ready to be deployed in the real world. Despite these challenges, well-constructed datasets can prove instrumental in encouraging models to learn specialised skills; datasets that enable agents to be trained across multiple environments can lead to more generalisable behaviour.

We further discuss the issues in evaluating \evlp~agents in real-world deployments, and we propose new assessment methods to address these challenges. We suggest that current evaluation paradigms should explicitly test the agents' generalisability across domains and tasks, as models need to be transferable to unseen environments and robust to model and environmental uncertainty. We also highlight the need for test beds with different (sub-)domain partitions, as current splits may not be representative of the variation in the intended real-world scenarios. Overall, our goal is to emphasise the importance of comprehensive evaluation paradigms for \evlp~agents and to propose new methods to assess their generalisability and robustness in real-world deployment.

\section{Conclusion}

In this extended abstract, we proposed a taxonomy for the field of \textit{\EVLP} (\evlp), which highlights: tasks, modelling approaches, learning paradigms, and evaluation settings; we provided a framework for discussing existing and future tasks, based on the skills required to solve them. We alluded to various learning paradigms, training strategies, and commonly-used optimisation objectives; and we considered different forms of evaluation---e.g., using datasets of expert demonstrations, simulators, and various task metrics. Finally, we focused on the challenges currently being tackled in the field, as well as those that still remain to be addressed. Specifically, we discuss issues that could prevent real-world deployment, such as a lack of generalisation, robustness, simulator realness, and lack of rich interaction; we highlight these as the most promising and fulfilling next directions to follow. 

%\section{Outlooks}

%\jf{We can consider two options -- (i) make this its own (sub-)section or (ii) integrate these topics back into Section \ref{sec:core_challenges}, above. I am more inclined to do the latter.}

%\begin{itemize}
%    \item Emphasize scenarios where agents \textit{must} learn through interaction and grounding -- interactive object perception, behaviour transfer \cite{tatiya2022transferring,tatiya2023cross}
%\end{itemize}

%\appendix

%\section*{Ethical Statement}

%There are no ethical issues.

\clearpage

\section*{Acknowledgments}

The authors thank Alessandro Oltramari, Yonatan Bisk, Eric Nyberg, and Louis-Philippe Morency for insightful discussions on the original journal article; the authors also thank Gyan Tatiya, Jimin Sun, Prasoon Varshney, Sahiti Yerramilli, and Jayant Tamarapalli for new/recent \evlp~collaborations. %This work was supported, in part, by a doctoral research fellowship from Bosch Research, by the U.S. Air Force Office of Scientific Research, under award number FA2386-17-1-4660, and by Jonathan's dual affiliation with the Human-Machine Collaboration group at Bosch Research Pittsburgh. The views expressed in this article do not necessarily represent those of the aforementioned entities.

%% The file named.bst is a bibliography style file for BibTeX 0.99c
\bibliographystyle{named}
\bibliography{ijcai23}

\end{document}